\documentclass{article}
\usepackage{iclr2027_conference}
\usepackage{iftex}
\ifPDFTeX
  \usepackage{times}
\else
  \usepackage{fontspec}
\fi

\usepackage{hyperref}
\hypersetup{
  colorlinks=true,
  citecolor=[RGB]{35,75,120},
  linkcolor=[RGB]{35,75,120},
  urlcolor=[RGB]{35,75,120}
}
\usepackage{url}
\usepackage{amsmath,amssymb,amsthm}
\usepackage{booktabs}
\usepackage{array}
\usepackage{graphicx}
\usepackage{microtype}
\usepackage{xspace}
\usepackage{cleveref}
\usepackage{algorithm}
\usepackage{algpseudocode}
\setcitestyle{numbers,square,comma}

\title{DeltaWAM: Change-centric Visual Foresight via Delta Tokens for Efficient World-Action Model}

\author{Tianyun Jiang$^{1}$ \And Wenrui Bao$^{2}$ \And Bingxin Xu$^{3}$ \And Yu Tian$^{2}$ \And Yuzhang Shang$^{2}$\thanks{Corresponding author.}
  \AND $^{1}$Fudan University \\
  $^{2}$University of Central Florida \\
  $^{3}$University of Southern California}

\iclrfinalcopy

\begin{document}

\maketitle
\lhead{Preprint}
\raggedbottom

\begin{abstract}
World-Action Models (WAMs) offer promising foresight for robotic manipulation,
yet efficient and dynamics-aware future representations haven't been fully established. Pixel-space WAMs use representations that serve for repainting entire scenes, which brings up computational overhead, spatial-temporal redundancy and waste of model capacities on task-irrelevant information. To resolve these inefficiencies, we start from an overlooked observation: consecutive frames in physical manipulation usually share substantial contexts and differ only in structured and low-dimensional ways (e.g., object displacements and end-effector motion). In essence, it is these dynamic changes, rather than sequences of entire scenes, that an action policy needs to anticipate. Grounded in this principle, we build \textbf{DeltaWAM}, a World-Action model that shifts the fundamental unit of future predictions to a delta token. Each delta token is a single 1D vector that encodes the changes of dense DINO features between consecutive frames, inherently capturing dynamic transitions while reducing redundancy from shared static context. Our DeltaWAM builds on DeltaWorld, a latent world model that is pretrained on large-scale videos and forecasts change-centric futures: given a history of observed delta tokens, it autoregressively predicts the future delta sequence, one token per frame. These predicted transitions, alongside the DINO features of current observations acting as the spatial anchor, are then passed to a flow-matching action expert that generates the action chunk. Our experiments show that by focusing merely on what will change in the future, delta tokens prove to be more dynamics-aware and action-forcing, providing a natural bridge from high-level semantics to low-level control. With just 256 GPU hours on 2 H100 GPUs and 0.725B total parameters, DeltaWAM achieves high success rates(92.8\%) on LIBERO while also demonstrating robust generalization under procedural perturbations on LIBERO-Pro. At inference the latency is 142.1 ms latency per action chunk and peak memory is only 3.86 GB.
Code is available at \url{https://github.com/deltawam/DeltaWAM}.
\end{abstract}

\section{Introduction}
\label{sec:intro}
World-Action Models (WAMs), which couple a predictive world model with action generation, have recently made rapid progress in robotic manipulation \citep{lv2025f1,ye2026dreamzero,kim2026cosmospolicy,li2026lingbotva,huang2026forewam,lei2026dele}. However, most existing WAMs synthesize full scenes at every future timestep rather than directly modeling dynamic changes. As a prime example, pixel-space WAMs typically adopt video generative backbones to imagine the future frame by frame \citep{hu2025vpp,zhu2025uwm,pai2025mimicvideo,lv2025f1,ye2026dreamzero,kim2026cosmospolicy,li2026lingbotva}. These models allocate substantial capacity to full pixel-level synthesis rather than distilling compact, action-relevant dynamics. Consequently, they spend much of their computation reconstructing task-irrelevant visual details that are largely repeated across frames, resulting in severe spatio-temporal redundancy. While latent WAMs circumvent pixel-level synthesis and operate in compressed feature spaces \citep{lyu2026lda,lin2026jepawam,chen2026lawam,lei2026dele,luo2026beingh07}, they inherit the same limitation. By forecasting absolute latent states frame by frame, they continue to model full scenes in the latent space rather than directly compressing dynamic transitions. This not only wastes models' representational capacity in maintaining static shared contexts, but also overshadows action-relevant dynamics that is important for low-level control.

In the physical world, \textbf{motion manifests as changes in state over time}. During action planning, the human visual and sensorimotor systems prioritize
task-relevant visual dynamics such as changes in object positions while attenuating static visual context such as the background details \citep{goodale1992visualpathways,desimone1995selective,treue1996attention,corbetta2002attention,whitney2003visualmotion,reynolds2004attention}. In the same vein, the predictive focus of a World-Action model must center on dynamic changes that the robot needs to realize, stripping away the redundant reconstruction of visual contexts shared across consecutive frames. 

Following this principle, we propose to represent the imagined future as a sequence of transitions rather than a sequence of scenes. We instantiate this idea with delta tokens, originally introduced in DeltaWorld \citep{kerssies2026deltatok} for efficient video world modeling. In DeltaWorld, a delta tokenizer compresses frame-to-frame DINOv3 \citep{simeoni2025dinov3} feature differences into a single compact delta token. We observe that this representation, though designed for video prediction, is particularly well suited to control. It discards the static context shared across frames and retains only what changes. Consequently, temporal dynamics are directly exposed as a compact control prior, allowing the policy to plan actions conditioned on explicit state transitions.

\begin{figure}[t]
    \centering
    \includegraphics[width=\textwidth]{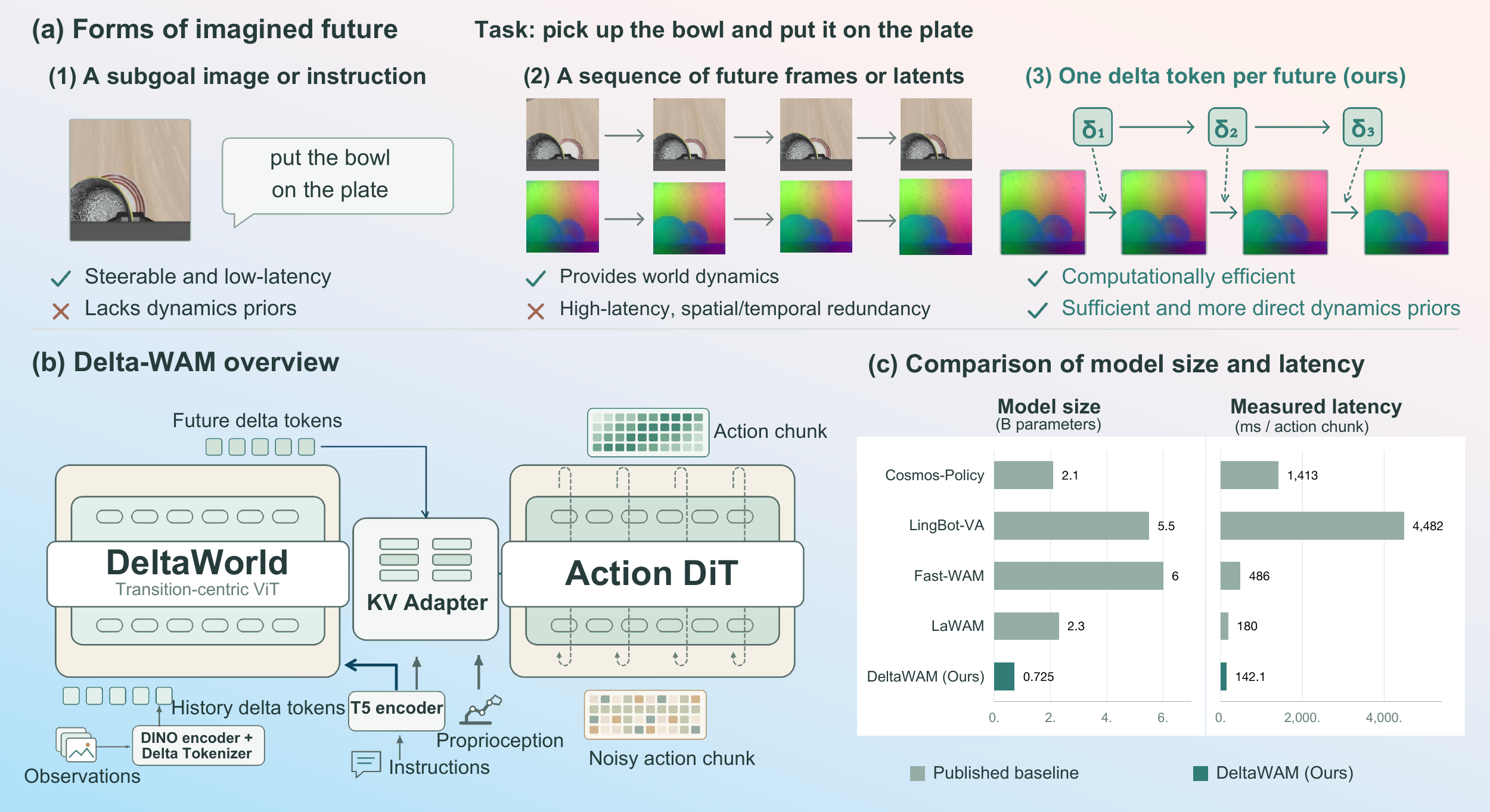}
    \vspace{-0.2in}
    \caption{\textbf{DeltaWAM overview.} (a) Existing policies represent an imagined future as a single subgoal or a sequence of dense frames or latent states, whereas DeltaWAM predicts one compact transition token per future step. (b) DeltaWAM couples a latent transition-centric world model DeltaWorld with an Action DiT, preserving efficient test-time visual foresight. (c) DeltaWAM is lightweight and has low latency compared with representative World-Action models.}
    \vspace{-0.2in}
    \label{fig:deltawam_overview}
\end{figure}

Turning DeltaWorld into a change-centric World-Action model raises two critical questions: how can the prediction itself be made more task-relevant, and how to feed the predicted transitions into the policy? Our DeltaWAM introduces a unified architecture to address both. The first question matters because we recognize that original DeltaWorld is inherently unconditional: it extrapolates futures based solely on observation history. In robotic manipulation, however, a single scene can unfold into multiple valid futures depending on the instruction. An unconditional rollout might therefore hallucinate a physically plausible future that misaligns with the task, ultimately misleading the downstream policy. To resolve this, we inject language instructions directly into the DeltaWorld Predictor via Gated AdaLN \citep{peebles2022dit,perez2018film}. An instruction embedding from a frozen text encoder modulates the scale, shift, and residual gates of every DeltaWorld block. For the second question, we pair DeltaWorld with a flow-matching Action DiT and condition it on two complementary signals. The first signal is the dense, patch-level DINO features of the current observation, which serve as a \textbf{spatial anchor} encoding precise static context (i.e., \textit{where} things are). The second signal is the sequence of predicted future delta tokens, which form a \textbf{dynamic subgoal} encapsulating motion intent (i.e., \textit{how} things should get moved). 

A controlled ablation validates our theory: providing the policy with full future DINO latents reduces the success rate on the standard LIBERO benchmark from 92.8\% to 79.0\% relative to conditioning on compressed delta tokens. On LIBERO-Pro, DeltaWAM achieves a 17.85\% average success rate under task perturbations, while the success rates of representative baselines are close to 0\%. This comparison further shows that instead of simply memorizing action trajectories for a given scene, DeltaWAM relies more on motion dynamics. To sum up, our contributions are three-fold:
\begin{itemize}
    \item We identify the spatio-temporal redundancy in existing WAMs and propose factorizing visual foresight into static visual anchors and dynamic transition tokens, fundamentally shifting world modeling from scene reconstruction to change prediction.
    \item We present DeltaWAM, a lightweight (0.725B parameters, 3.86GB peak memory at inference) and computationally efficient World-Action model.
    \item Our experiments substantiate that compressed delta tokens provide a superior, action-forcing signal compared to full state predictions.
\end{itemize}

\section{Related Work}
\label{sec:related}



\subsection{World-Action Models}

Vision-Language-Action (VLA) models build on large-scale vision-language pretraining, which connects language instructions with target objects and task structure \citep{brohan2023rt2,kim2024openvla,black2024pi0,black2025pi05}. This semantic prior tells the robot what to do, but lacks dynamics priors of how the environment will evolve during interaction \citep{pai2025mimicvideo,ye2026dreamzero,chen2026lawam}. 
World-Action Models (WAMs) address this gap by coupling action generation with future prediction\citep{hu2025vpp,zhu2025uwm,pai2025mimicvideo,lv2025f1,ye2026dreamzero,kim2026cosmospolicy}.
Video-based WAMs leverage rich dynamics priors inherited from web-scale video pretraining to generate future frame sequences \citep{hu2025vpp,zhu2025uwm,pai2025mimicvideo,ye2026dreamzero,kim2026cosmospolicy,li2026lingbotva}.
However, their reliance on iterative pixel-level video diffusion introduces severe computational latency, creating a fundamental deployment bottleneck for real-time closed-loop control \citep{yuan2026fastwam,ye2026gigaworld,chen2026lawam,li2026lightwam}. To mitigate this overhead, efficiency-oriented architectures like Fast-WAM and Giga-World Policy either relegate world prediction to an auxiliary loss or formulate action-centered generation \citep{yuan2026fastwam,ye2026gigaworld}.
While faster, these compromises weaken the functional role of predicted dynamics as explicit visual subgoals at test time.
To preserve explicit test-time foresight without the burden of pixel rendering, recent Latent WAMs shift prediction to representation spaces not optimized for pixel reconstruction \citep{lyu2026lda,lin2026jepawam,chen2026lawam,su2026wog,luo2026beingh07,yang2026lilawam,huang2026forewam,lei2026dele}.
For instance, LDA-1B models future states in a DINO feature space and jointly denoises future representations and action chunks~\citep{lyu2026lda}. JEPA-WAM predicts future V-JEPA latents via a language backbone, yet utilizes
latent predictions purely as auxiliary conditioning during training rather than execution \citep{lin2026jepawam}.
LaWAM predicts chunk-level DINO visual subgoals, but requires additional latent action distillation to guide world prediction \citep{chen2026lawam}. Fundamentally, these models still predict latent for full scenes at each future timestep without extracting dynamic changes.

By contrast, our DeltaWAM also operates in DINO latent space but compresses the DINO differences between consecutive frames into a single compact delta token. Rather than iterative denoising, each delta token's generation requires only a single forward pass, which is more computational efficient. By decoupling future dynamic prediction from raw pixel synthesis and exposing the dynamic transitions directly to the policy, DeltaWAM provides stronger visual foresight for low-level control.

\subsection{Latent Representations for World Modeling}
Effective physical foresight in embodied decision-making relies heavily on the
choice of world model state space
\citep{ha2018worldmodels,hafner2019planet,hansen2024tdmpc2,zhou2024dinowm,assran2025vjepa2}.
Early visual world models typically operated in pixel space or compressed variational autoencoder (VAE) latents designed primarily for image reconstruction
\citep{finn2016video,ha2018worldmodels,hafner2019planet,hafner2020dreamer}.
However, raw visual representations exhibit severe spatial and temporal redundancy, and generative capacity is wasted on reconstructing task-irrelevant background details \citep{hu2025vpp,kerssies2026deltatok,chen2026lawam,luo2026beingh07}.

To overcome these representation limits, self-supervised visual foundation models (VFMs) such as DINO capture high-level semantic layout and spatial structure without serving for pixel reconstruction \citep{oquab2023dinov2,simeoni2025dinov3}.
Building upon DINOv3 features, DeltaWorld compresses frame-to-frame feature differences into compact, one-dimensional delta tokens~\citep{kerssies2026deltatok} for more effficient world modeling. Our work leverages this transition-centric latent paradigm, demonstrating that autoregressive prediction of low-dimensional delta tokens provides a highly
efficient, action-forcing dynamic prior for downstream low-level control.

\begin{figure}[t]
    \centering
    \includegraphics[width=\textwidth]{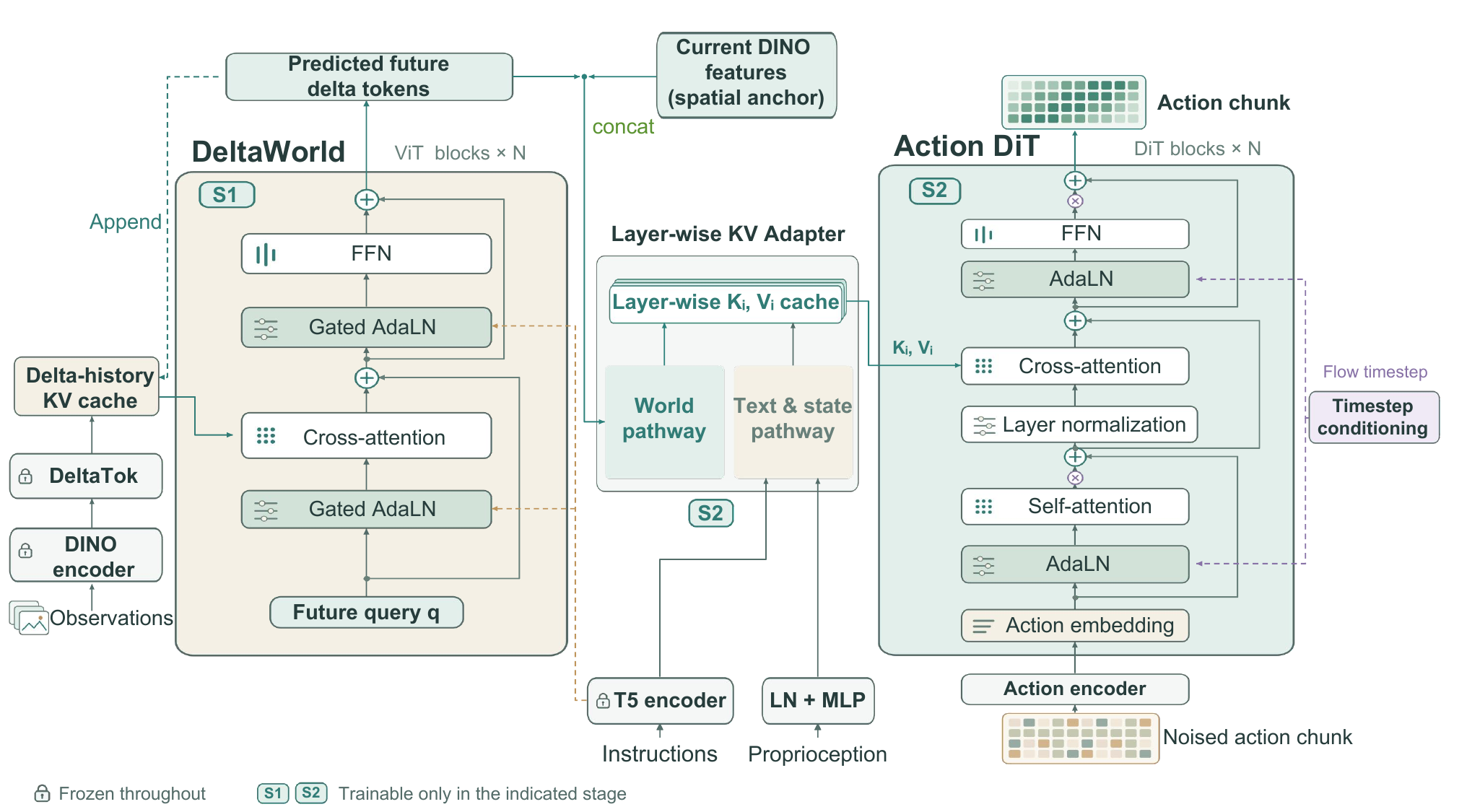}
    \caption{\textbf{Detailed DeltaWAM architecture.} DeltaWAM couples the change-centric DeltaWorld with a flow-matching Action DiT. The pretrained DeltaWorld is first finetuned on demonstrations; then Action DiT is trained on robot data with the finetuned DeltaWorld frozen. DeltaWorld first rolls out predicted delta tokens autoregressively. Afterwards, both current DINO latents (the spatial anchor) and predicted delta tokens (the dynamic subgoal) are fed to Action DiT.}
    \label{fig:deltawam_architecture}
\end{figure}

\section{METHOD}
In this section, we present DeltaWAM, a framework that integrates transition-centric delta token prediction with low-level action generation. We first introduce the concept of delta tokens from DeltaWorld in Section 3.1. We then present task-conditioned world modeling in Section 3.2. Finally, in Section 3.3, we describe the training paradigms for the lightweight and computationally efficient DeltaWAM.
\subsection{Preliminary}
\paragraph{Delta Tokenization}
    Given a multi-view visual sequence $(O_1, \dots, O_t)$, a frozen visual foundation backbone DINOv3 \citep{simeoni2025dinov3} extracts dense patch-level spatial feature maps $X_\tau = \phi(O_\tau) \in \mathbb{R}^{V \times N \times D_v}$ for each frame $\tau$, where $N$ is the number of spatial patches per view and $D_v$ is the feature dimension. In DeltaWorld \citep{kerssies2026deltatok}, the frame-level delta tokenizer is constructed based on a continuous autoencoder
design. To initialize absolute visual grounding, a synthetic reference frame $O_\emptyset = \mathbf{0}$ with features $X_\emptyset = \phi(O_\emptyset)$ is prepended. The tokenizer encoder $g$ compresses each pair of consecutive spatial feature maps into a 1D sequence of delta tokens $Z_{1:t} = (z_1, \dots, z_t)$, where $z_\tau \in \mathbb{R}^{V \times 1 \times D_v}$, $z_1 = g(X_\emptyset, X_1, z_{\text{init}})$, and $z_\tau = g(X_{\tau-1}, X_\tau, z_{\text{init}})$ for $\tau > 1$.
Here, $z_{\text{init}} \in \mathbb{R}^{1 \times D_v}$ is a learnable aggregation token prepended to the paired spatial feature maps before being fed into $g$. The tokenizer decoder $h$ reconstructs the current feature map by transforming the previous features according to the delta token.
The delta tokenizer is then trained using a
reconstruction loss between the original and reconstructed feature maps.
This objective encourages the delta token $z_{\tau}$ to serve as a compact transition representation between consecutive frames. Note that a timestamp sampling procedure can be employed for training pairs of $(X_{\tau-1},X_{\tau})$, which means that a single delta token can represent transitions ranging from nearly static scenes to substantial scene changes. 

\subsection{Task-conditioned World Modeling with Delta Tokens}
Let $Z_{\text{obs}}=Z_{t-T_c+1:t}$ denote the observed history context of delta tokens, with one delta token per camera view at each timestep. DeltaWorld generates each timestep's delta tokens autoregressively. At rollout step $j$, it predicts the next tokens for all views in parallel, conditioned on the history context and its previous predictions:
\begin{equation}
    \hat{z}_{t+j}^{1:V}
    = P_\psi\!\left(q_{t+j}, Z_{\text{obs}}, \hat{Z}_{t+1:t+j-1}, c\right),
    \qquad j=1,\dots,L_w,
\end{equation}
where $q_{t+j}$ is a stochastic query sampled from a Gaussian distribution and shared across views, and $c$ is the task embedding. We start from DeltaWorld pretrained on Kinetics-700, a large-scale video dataset containing diverse human activities and object interactions \citep{carreira2019kinetics700}. To inject high-level semantics, we encode the task instruction with a frozen T5 encoder \citep{raffel2020t5}, project its token features to the predictor dimension, and obtain the task embedding $c$. The resulting embedding modulates each attention and MLP branch through Gated Adaptive Layer Normalization (Gated AdaLN) \citep{perez2018film,peebles2022dit}. For a branch input $h$ and transformation $F$,
\begin{equation}
    \tilde{h} = \left(1 + \gamma(c)\right) \odot \operatorname{LN}(h) + \beta(c), \qquad
    h_{\mathrm{out}} = h + \left(1 + \eta(c)\right) \odot F(\tilde{h}),
\end{equation}
where $\gamma$, $\beta$, and $\eta$ are language-dependent scale, shift, and residual-gate parameters. Crucially, their output projections are zero-initialized---this ensures that fine-tuning smoothly leverages the pretrained dynamics prior, allowing language to \textit{steer} the learned physics rather than \textit{overwrite} it.

\begin{figure}[H]
    \centering
    \small
    \includegraphics[width=\linewidth]{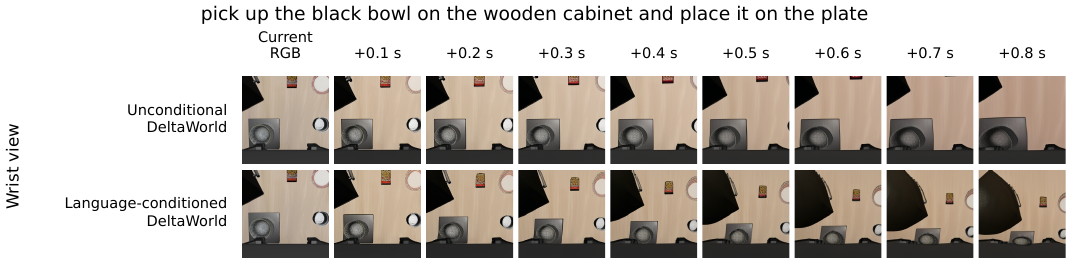}
    \caption{\textbf{Language steers predicted futures toward the task goal.} Wrist-view rollouts for LIBERO-Spatial task 9, which asks the robot to pick up the black bowl on the wooden cabinet and place it on the plate. In DeltaWorld's unconditional predictions, the robot approaches the stove instead of the cabinet. This is misleading since the bowl is on the cabinet.}
    \label{fig:task9_language_prediction}
\end{figure}

\subsection{Training Paradigm}
\subsubsection{Stage 1: BoM Training for Task-Conditioned DeltaWorld}
Following the original DeltaWorld training recipe \citep{kerssies2026deltatok}, we adopt a Best-of-Many (BoM) training paradigm \citep{bhattacharyya2018bom}. At each supervised position $\tau$, $K$ Gaussian queries are sampled and go through the forward pass in parallel. For each candidate $k$, the same query is shared across different camera viewpoints, allowing the
DeltaWorld predictor $P_\psi$ to predict a coherent multi-view hypothesis:
\begin{equation}
    \hat{z}_{\tau}^{(k),1:V}
    =
    P_\psi\left(q_{\tau}^{(k)},Z_{<\tau},c\right),
    \qquad k=1,\ldots,K.
\end{equation}
Here causal teacher forcing on the clean target sequence $Z_{1:T}$ is applied. For every supervised position $\tau \in \{2, \dots, T\}$, the predictor attends only to the ground-truth prefix $Z_{<\tau}$ through a causal mask, which prevents access to the current and future targets. Let $\rho_{\mathrm{SL1}}$ denote the Smooth-L1 prediction loss of predicted delta tokens averaged over all the camera viewpoints. Using this loss, we can get the candidate that is closest to the ground-truth future:
\begin{equation}
    \mathcal{E}_{\tau}^{(k)}
    =
    \frac{1}{V}
    \sum_{v=1}^{V}
    \rho_{\mathrm{SL1}}
    \left(z_{\tau}^{v},\hat{z}_{\tau}^{(k),v}\right),
    \qquad
    k_{\tau}^{*}
    =
    \arg\min_{k\in\{1,\ldots,K\}}
    \mathcal{E}_{\tau}^{(k)}.
\end{equation}
Then at each position only the selected candidate gets supervised by minimizing the training objective $\mathcal{L}_{\mathrm{world}} = \frac{1}{T-1}\sum_{\tau=2}^{T}\mathcal{E}_{\tau}^{(k_{\tau}^{*})}$.
In other words, gradients are propagated only through the selected candidate.  Because each position is trained under teacher forcing with its own Gaussian queries, the selected candidate index is not required to remain fixed across the sequence.

The BoM training paradigm brings several advantages. Firstly, it allows DeltaWorld to be generative rather than discriminative. When multiple future outcomes are
plausible, a regression objective tends to drive the model toward an averaged prediction that may not correspond to any realistic outcomes. Also, compared with diffusion-based generative models that involve iterative denoising, this is more computationally efficient. Another main point is that the downstream policies can be multimodal---there are multiple ways to finish a single task. The BoM training objective leaves space for imagining diverse visual subgoals with a unified motion intent.

\subsubsection{Stage 2: Delta-Conditioned Policy Training}
We freeze the task-finetuned DeltaWorld predictor and train Action DiT together with its KV adapter. DeltaWorld first autoregressively samples $K$ candidate future trajectories. An "oracle" candidate branch $k^\dagger$ is identified by the minimal trajectory error:
\begin{equation}
    k^\dagger = \arg\min_{k \in \{1, \dots, K\}} \sum_{j=1}^{L_w} \sum_{v=1}^V \| z_{t+j}^v - \hat{z}_{t+j}^{(k), v} \|_2^2.
\end{equation}
 During training, Action DiT conditions on the "oracle" branch $k^\dagger$ with probability $p_{\text{oracle}}$ and on a randomly selected branch from $\{1, \dots, K\}$ with probability $1 - p_{\text{oracle}}$. This design makes sure that Action DiT is more likely to condition on the "oracle" branch by setting a reasonable $p_{\text{oracle}}$. The prediction quality of non-oracle branches is unstable; therefore, we cannot sample all branches uniformly.
 For a ground-truth action chunk $a$, Gaussian noise $\epsilon\sim\mathcal N(0,I)$, and flow time $\sigma$, let $a_\sigma=(1-\sigma)a+\sigma\epsilon$. Given the selected conditioning context $\mathcal C$, we minimize:
\begin{equation}
    \mathcal L_{\mathrm{action}}=\mathbb E_{a,\epsilon,\sigma}\!\left[w(\sigma)\left\|v_\theta(a_\sigma,\sigma;\mathcal C)-(\epsilon-a)\right\|_F^2\right]
\end{equation}
where $v_\theta$ is Action DiT's predicted velocity and $w(\sigma)$ is the flow-time weight. At inference, a future trajectory is sampled from the DeltaWorld predictor for generation of one action chunk. 
\vspace{-7pt}
\begin{figure}[H]
    \centering
    \includegraphics[width=0.84\linewidth,trim=0 0 0 52,clip]{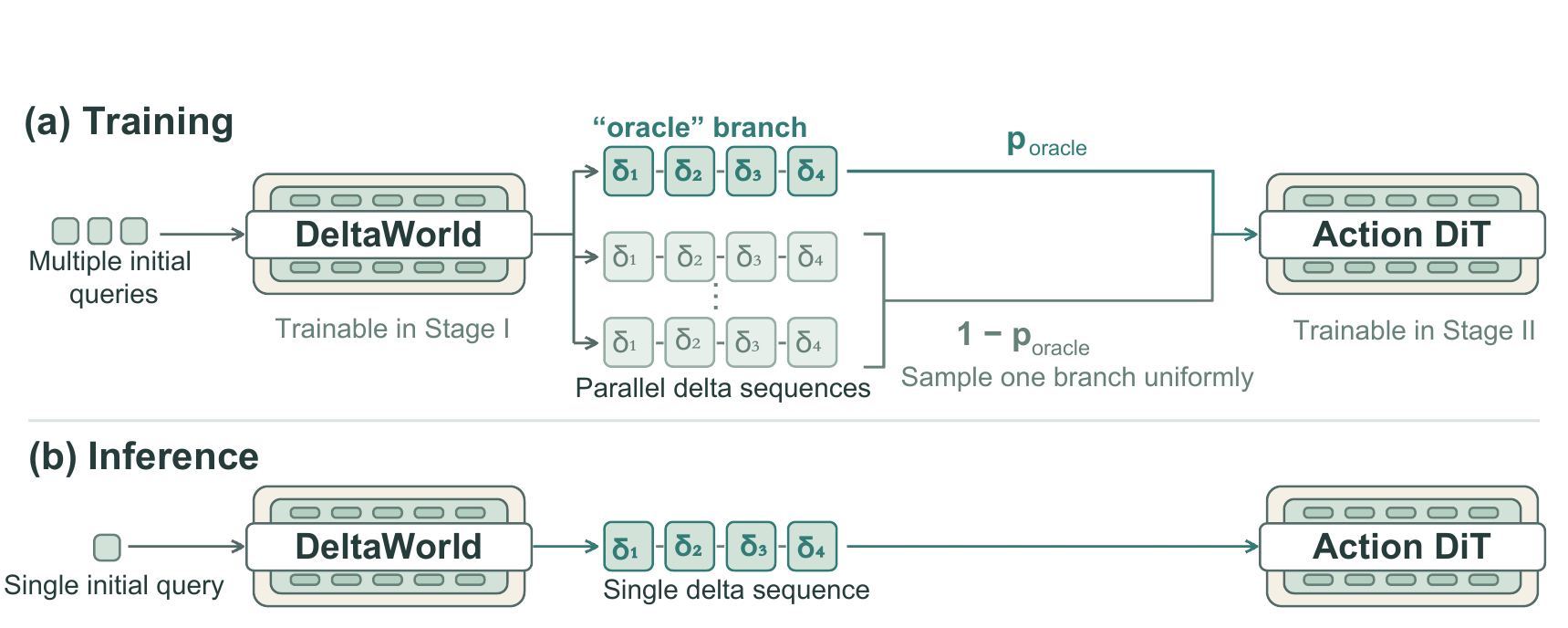}
    \caption{\textbf{DeltaWAM training and inference.} During the second training stage, Action DiT learns from oracle-selected or randomly sampled future trajectories produced by frozen DeltaWorld. At inference, one sampled future trajectory conditions action generation.}
    \label{fig:training_inference}
\end{figure}
\section{EXPERIMENTS}
\setlength{\textfloatsep}{10pt plus 2pt minus 2pt}
\setlength{\floatsep}{9pt plus 2pt minus 2pt}
\setlength{\intextsep}{9pt plus 2pt minus 2pt}
\setlength{\abovecaptionskip}{5pt}
\setlength{\belowcaptionskip}{2pt}
\renewcommand{\topfraction}{0.95}
\renewcommand{\bottomfraction}{0.90}
\renewcommand{\textfraction}{0.05}
\renewcommand{\floatpagefraction}{0.82}
\setcounter{topnumber}{3}
\setcounter{bottomnumber}{2}
\setcounter{totalnumber}{5}
\makeatletter
\setlength{\@fptop}{0pt}
\setlength{\@fpsep}{9pt plus 1fil}
\setlength{\@fpbot}{0pt plus 1fil}
\makeatother
We aim to evaluate whether compact delta-token foresight enables effective manipulation at a low inference cost, and whether it exhibits representational superiority in providing dynamics priors for low-level control. Our experiments examine closed-loop task success and latency on LIBERO, robustness to visual and procedural distribution shifts, and the contribution of predicted delta tokens for task execution. We also conduct the real-world evaluation on a PiPER robotic arm.
\subsection{Experimental Setup}
\paragraph{Benchmarks.}
LIBERO is a simulated benchmark for various manipulation tasks, and we evaluate across its four suites: Spatial, Object, Goal, and LIBERO-10 (Long) \citep{liu2023libero}. LIBERO-Pro tests procedural robustness beyond the standard tasks, and we use its task-perturbation dimension, which redefines task logic and target states\citep{zhou2025liberopro}.
\paragraph{Training and Evaluation Protocol}
For simulation, we first fine-tune the pretrained DeltaWorld predictor on LIBERO demonstrations, and then freeze it to train suite-specialized policies. The entire training pipeline consumes merely 256 GPU hours on two H100 GPUs, dramatically lowering the training barrier compared to prior WAMs. At test time, DeltaWorld samples a future delta sequence and the action expert generates a 16-step action chunk with ten Euler steps, executing eight actions before replanning. Following the standard protocol, on LIBERO and LIBERO-Pro we evaluate each task 50 times and run the test with 3 different random seeds. Architecture, optimization, and evaluation details appear in Appendix~\ref{app:implementation}.
\paragraph{Real-World Manipulation Setup}
For the real-world evaluation, we deploy DeltaWAM on an AgileX PiPER six-DoF robotic arm. We use a 150-episode pick-and-place manipulation dataset collected at 30 Hz, with RGB observations from front and wrist cameras together with 14-dimensional robot states. The policy predicts a seven-dimensional action consisting of six absolute joint-position targets and a gripper command.
\subsection{Main Performance and Efficiency}
Table~\ref{tab:libero_main} jointly compares success rates, model size and model-only query latency across the four standard LIBERO suites. Latency results mainly come from LaWAM \citep{chen2026lawam} and our measurement strictly follows its evaluation protocol. DeltaWAM achieves an average success
rate of 92.8\%. In both LIBERO-Spatial and LIBERO-Long, the failure cases mainly stem from instability during precision contact---the robot successfully puts the item on the plate, yet the placement is off-center. This is an expected consequence of the architecture, since highly compressed delta tokens may filter out the fine-grained spatial nuances necessary for precise alignment.

\begin{table}[tbp]
\centering
\small
\renewcommand{\arraystretch}{0.96}
\caption{Standard LIBERO success rate (\%), model size, and model-only latency
per action-chunk query. Emb. PT denotes embodied action pretraining. Baseline sizes
and latency values follow their source papers and the comparison protocol of
LaWAM \citep{chen2026lawam}.}
\label{tab:libero_main}
\resizebox{\textwidth}{!}{%
\begin{tabular}{lcccccccc}
\toprule
\textbf{Method} & \textbf{Size} & \textbf{Emb. PT} & \textbf{Latency (ms)} & \textbf{Spatial} & \textbf{Object} & \textbf{Goal} & \textbf{Long} & \textbf{Average} \\
\midrule
\multicolumn{9}{l}{\textit{Reactive Policies}} \\
Diffusion Policy~\citep{chi2023diffusionpolicy} & 263M & $\times$ & -- & 78.3 & 92.5 & 68.3 & 50.5 & 72.4 \\
OpenVLA~\citep{kim2024openvla} & 7B & $\checkmark$ & -- & 84.7 & 88.4 & 79.2 & 53.7 & 76.5 \\
$\pi_0$~\citep{black2024pi0} & 3.5B & $\checkmark$ & 220 & 98.0 & 96.8 & 94.4 & 88.4 & 94.4 \\
$\pi_{0.5}$~\citep{black2025pi05} & 3.5B & $\checkmark$ & 220 & 98.8 & 98.2 & 98.0 & 92.4 & 96.9 \\
GR00T-N1.6~\citep{nvidia2025grootn16} & 3.3B & $\checkmark$ & 259 & 97.7 & 98.5 & 97.5 & 94.4 & 97.0 \\
\midrule
\multicolumn{9}{l}{\textit{World-action models}} \\
F1~\citep{lv2025f1} & 4B & $\checkmark$ & 399 & 98.2 & 97.8 & 95.4 & 91.3 & 95.7 \\
Fast-WAM~\citep{yuan2026fastwam} & 6B & $\times$ & 486 & 98.2 & \textbf{100.0} & 97.0 & 95.2 & 97.6 \\
Cosmos-Policy~\citep{kim2026cosmospolicy} & 2.1B & $\times$ & 1413 & 98.1 & \textbf{100.0} & 98.2 & 97.6 & 98.5 \\
Motus~\citep{bi2025motus} & 8B & $\checkmark$ & 3231 & 96.8 & 99.8 & 96.6 & 97.6 & 97.7 \\
LingBot-VA~\citep{li2026lingbotva} & 5.5B & $\checkmark$ & 4482 & 98.5 & 99.6 & 97.2 & \textbf{98.5} & 98.5 \\
LaWAM~\citep{chen2026lawam} & 2.3B & $\checkmark$ & 187 & \textbf{99.4} & 99.6 & \textbf{98.4} & 97.0 & \textbf{98.6} \\
\midrule
\textbf{DeltaWAM (Ours)} & \textbf{0.725B} & $\boldsymbol{\times}$ & \textbf{142.1} & 92.6 & 97.0 & 95.8 & 85.8 & 92.8 \\
\bottomrule
\end{tabular}
}
\end{table}

DeltaWAM improves over small reactive policies but remains below the strongest pretrained VLA and WAM baselines. Its principal advantage in this comparison is the combination of manipulation performance and a much smaller computational footprint, rather than the highest success rate. Note that DeltaWAM is trained on LIBERO from scratch, which puts it at a disadvantage compared with policies pretrained on enormous robot datasets. Compared with other baseline models, our model is extremely lightweight with low inference costs. DeltaWAM uses 0.725B parameters in total, including 120.3M trainable parameters, and achieves an average latency of 142.1~ms per chunk after querying 100 times on an A100. The peak GPU memory at inference is about 3.86 GB and the estimated cost is 0.584 TFLOPs per action.

\subsection{Procedural Robustness of DeltaWAM}
Task perturbations in the LIBERO-PRO benchmark alter task-relevant factors such as the target object, target receptacle, spatial relation, or instruction semantics, requiring the policy to correctly interpret the changed goal rather than simply reproduce behaviors associated with the original task. They therefore evaluate whether a model can generalize its learned visuomotor skills to novel task configurations and remain responsive to changes in task intent. As shown in Table~\ref{tab:libero_pro}, many baselines' success rates under task perturbations are close to 0\%, which stands in sharp contrast with their high success rates on the original LIBERO tasks. This indicates that very likely these models' success stems from mechanical memorization of training scenarios rather than genuine acquisition of transferable task-solving strategies.

\begin{figure}[H]
    \centering
    \begin{minipage}[t]{0.44\linewidth}
        \centering
        \includegraphics[width=\linewidth]{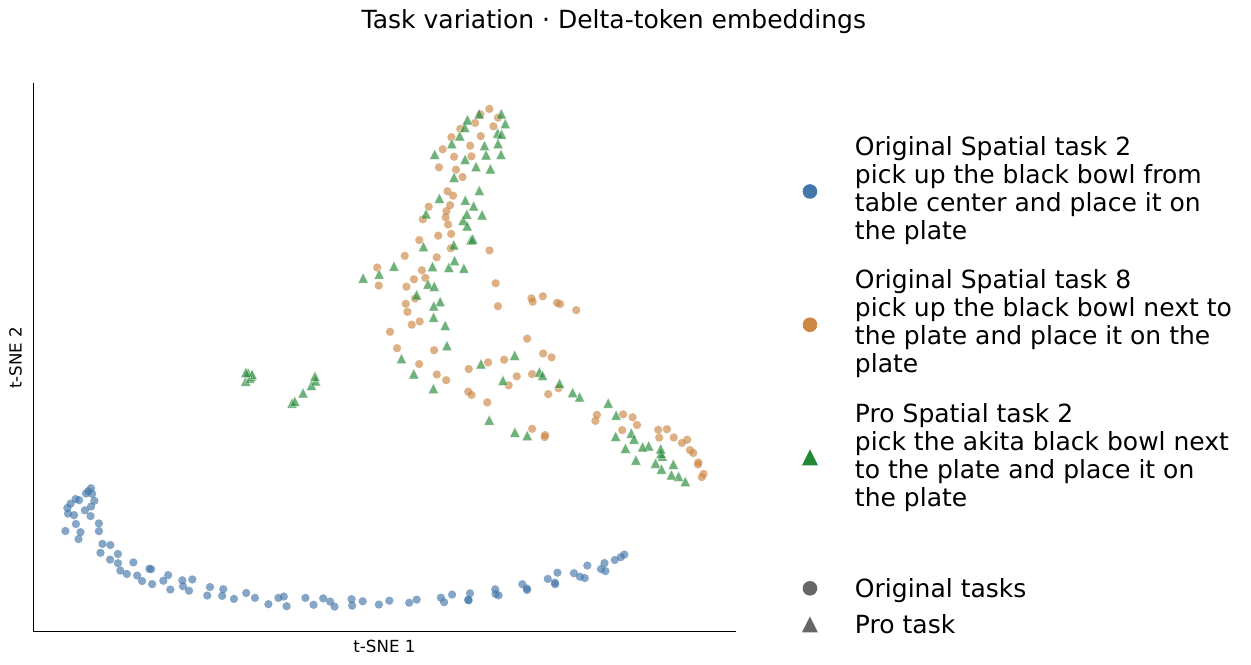}
        (a) LIBERO-Pro: Spatial task 2
    \end{minipage}
    \hspace{0.03\linewidth}
    \begin{minipage}[t]{0.44\linewidth}
        \centering
        \includegraphics[width=\linewidth]{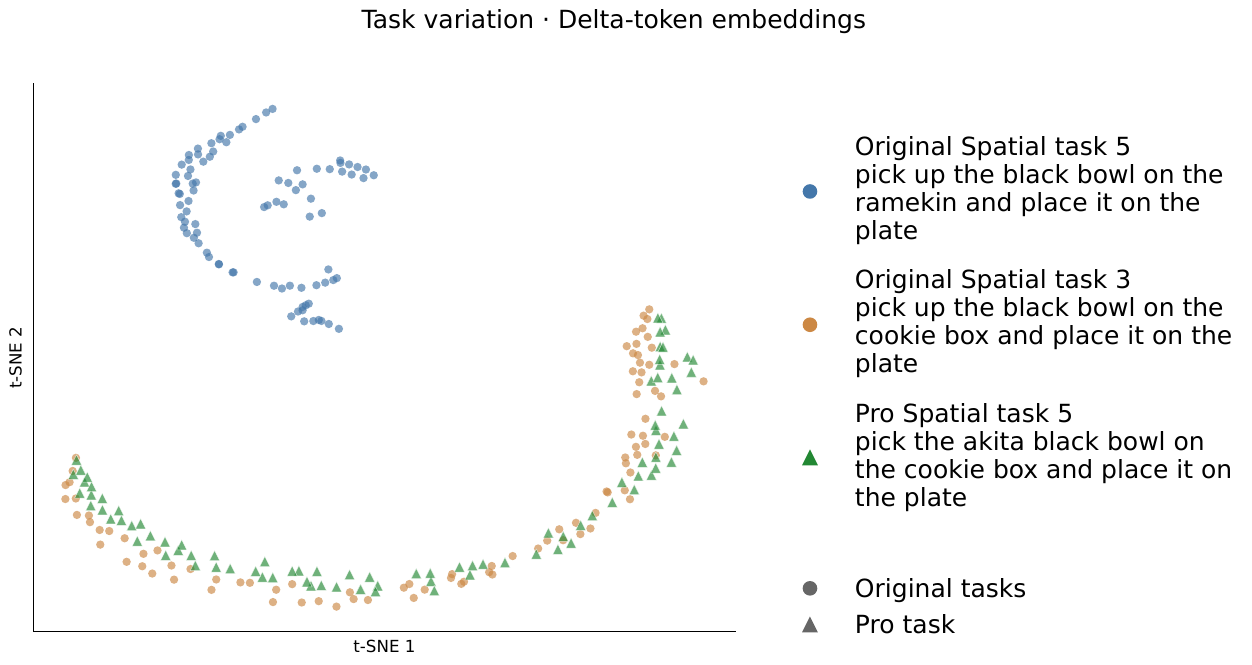}
        (b) LIBERO-Pro: Spatial task 5
    \end{minipage}
    \caption{\textbf{Delta-token structure under LIBERO-Pro task variations.} The two t-SNE projections compare predicted delta tokens from a perturbed task (green triangles) with tokens from its nominal original task (blue) and an original task whose task requirements match the perturbation (orange).}
    \label{fig:libero_pro_delta_tsne}
\end{figure}

\begin{table}[H]
\centering
\small
\renewcommand{\arraystretch}{0.96}
\caption{Success rate (\%) under the LIBERO-Pro task perturbation, broken down by the four standard task suites. Official benchmark baselines follow \citet{zhou2025liberopro}. We evaluated the results for Fast-WAM and Lingbot-VA ourselves.}
\label{tab:libero_pro}
\begin{tabular}{lccccc}
\toprule
\textbf{Method} & \textbf{Spatial} & \textbf{Object} & \textbf{Goal} & \textbf{Long} & \textbf{Average} \\
\midrule
OpenVLA~\citep{kim2024openvla} & 0.0 & 0.0 & 0.0 & 0.0 & 0.0 \\
$\pi_0$~\citep{black2024pi0} & 0.0 & 0.0 & 0.0 & 0.0 & 0.0 \\
$\pi_{0.5}$~\citep{black2025pi05} & 1.0 & 1.0 & 0.0 & 1.0 & 0.75 \\
MolmoAct~\citep{lee2025molmoact} & 0.0 & 0.0 & 0.0 & 6.0 & 1.5 \\
NORA~\citep{hung2025nora} & 0.0 & 0.0 & 0.0 & 0.0 & 0.0 \\
x-VLA~\citep{zheng2025xvla} & 0.0 & 8.0 & 9.0 & \textbf{10.0} & 6.75 \\
Fast-WAM~\citep{yuan2026fastwam} & 8.0 & 6.6 & 6.2 & 7.6 & 7.1 \\
LingBot-VA~\citep{li2026lingbotva} & 2.0 & 1.0 & 1.0 & 0.0 & 1.0 \\
\midrule
\textbf{DeltaWAM (Ours)} & \textbf{42.2} & \textbf{10.2} & \textbf{9.2} & 9.8 & \textbf{17.85} \\
\bottomrule
\end{tabular}
\end{table}
In comparison, our DeltaWAM is more robust to task perturbations, especially on LIBERO-Spatial. Figure~\ref{fig:libero_pro_delta_tsne}(a,b) provides a qualitative view of how the
predicted representation responds to task variations. In both examples, the delta tokens of the perturbed task align more closely with those of a training task that shares the same underlying manipulation requirement, despite differences in task configuration and resulting motion trajectories, than with those of its nominal source task. This pattern demonstrates DeltaWAM's procedural robustness: it sticks to the task-relevant dynamics, rather than trajectories tied with a specific scene layout. This robustness arises from the representation itself: by filtering out shared context and merely focusing on dynamic changes, delta tokens remain more invariant across different task configurations that share the same underlying manipulation requirement. The evaluation results on LIBERO-PLUS in Appendix~\ref{app:libero_plus} further prove this.

\subsection{Representation Benefits: Ablate The Role of Delta Tokens}
\label{sec:representation_ablation}
\begin{figure}[H]
\centering
\begin{minipage}[t]{0.49\linewidth}
    \centering
    \includegraphics[width=\linewidth]{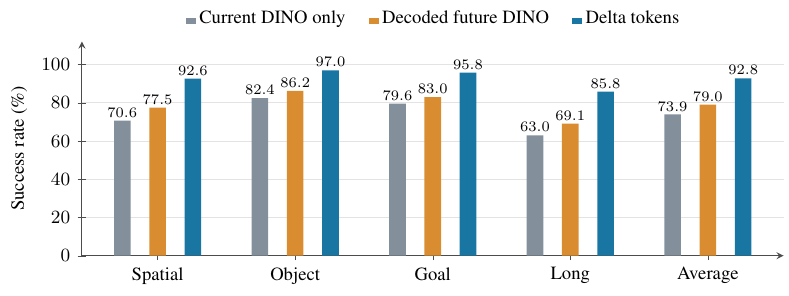}
    (a) Performance Comparison 
\end{minipage}
\hfill
\begin{minipage}[t]{0.49\linewidth}
    \centering
    \includegraphics[width=\linewidth]{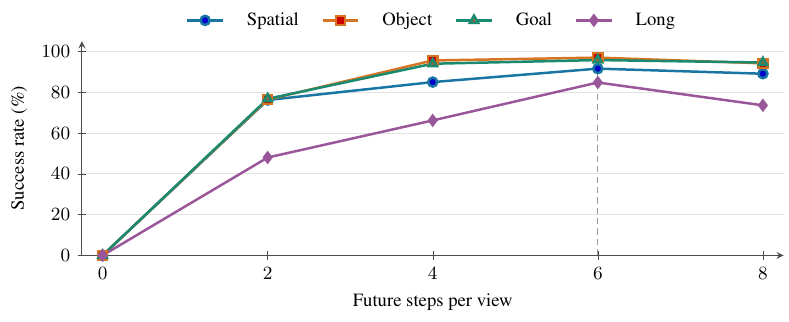}
    (b) Test-time Ablation Results
\end{minipage}
\caption{\textbf{Training and Test Ablations.} (a) Success rates of three models across the four task suites on LIBERO. Current DINO only, decoded future DINO, and delta tokens use 2,048, 14,336, and 2,060 visual tokens in total for one action chunk, respectively. (b) Comparison of success rates on LIBERO as the number of predicted future steps per camera view varies. Each step contributes one delta token per view, and the dashed line marks the default six-step horizon.}
\label{fig:delta_ablation_panels}
\end{figure}

To fairly evaluate the importance of predicted delta tokens for task performance, we construct two model variants. The \emph{current-DINO-only} variant bypasses the
DeltaWorld rollout and conditions the policy only on DINO patches of the current observation, while the \emph{decoded-future-DINO} variant retains DeltaWorld's predictions but decodes every future delta token back into the DINO latent. The two model variants receive the same training and inference configuration as the original DeltaWAM, and the only difference is what the action expert conditions on. As shown in Figure~\ref{fig:delta_ablation_panels}(a), performance drops by 18.9 and 13.8 percentage points, respectively. The \emph{current-DINO-only} variant has the lowest average success rate of 73.9\%, suggesting the necessity of future predictions. The comparison between the \emph{decoded-future-DINO} variant and the original DeltaWAM indicates that delta tokens are more action-forcing by exposing dynamic changes. Compared with the \emph{decoded-future-DINO} variant, DeltaWAM reduces future tokens by roughly $1{,}024\times$. Therefore we do not need to model the full state at each future timestep, and a change-centric representation is actually more compact and more beneficial for policy learning.

To assess how reliant DeltaWAM is on predicted delta tokens, we further conduct a test-time ablation. During training, the future step is set to six, which corresponds to twelve future action steps owing to temporal subsampling. Figure~\ref{fig:delta_ablation_panels}(b) compares success rates on the standard LIBERO benchmark by varying the number of predicted delta tokens. Removing all future tokens reduces success rates to 0\% on all four suites, showing that the trained policies rely strongly on predicted futures. We observe that in the absence of predictive guidance, the policy degenerates into aimless reaching and grasping behaviors. With only two future steps per view, success recovers to 76.2\%, 76.4\%, 76.8\%, and 48.0\% on Spatial, Object, Goal, and Long. Longer foresight is not uniformly beneficial---there is a trade-off between additional future context and prediction error. The analysis of accumulating errors in DeltaWorld's predictions and the ablation study of training policies with the full-horizon future are provided in the appendix. 

\begin{figure}[H]
\centering
\includegraphics[width=\linewidth]{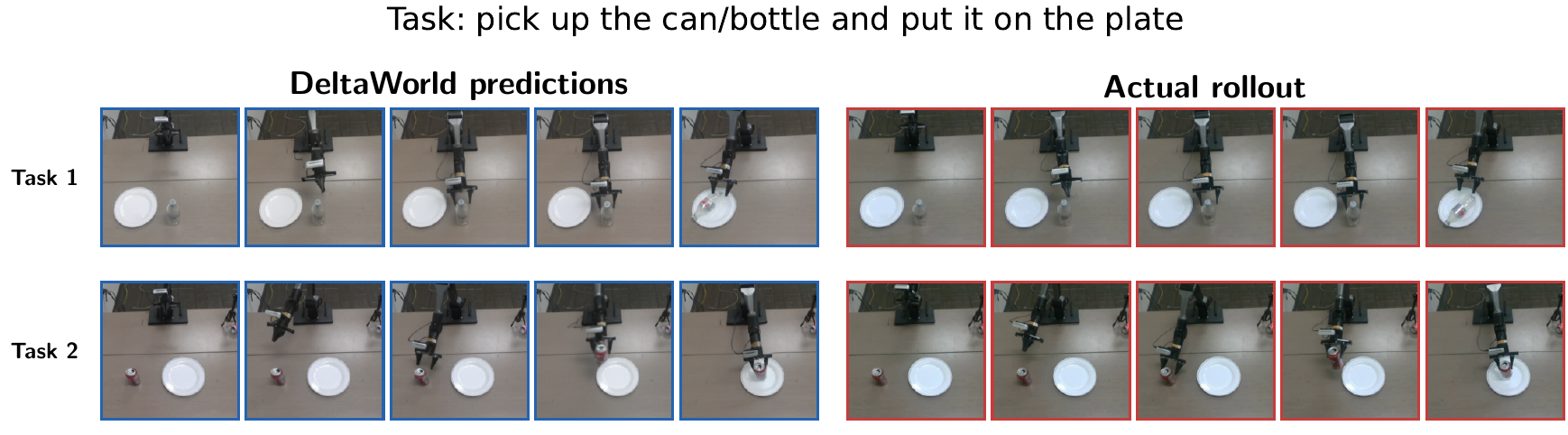}
\caption{\textbf{Real-world deployment} We conduct real-world experiments on a Piper robotic arm. The task is to pick up the bottle/can and place it on the plate. DeltaWorld first predicts the delta tokens, then the Action DiT predicts the action chunk for execution. The frozen delta tokenizer decoder and DINO-to-RGB decoder is applied for DeltaWorld's predictions.}
\label{fig:real_robot_phases}
\end{figure}

\section{Conclusion}
\label{sec:conclusion}
In this work, we propose DeltaWAM, a fast and lightweight World-Action model that shifts the prediction unit to a single compact delta token. Experiments on LIBERO show that this change-centric representation supports strong manipulation performance with low latency and provides more effective dynamics priors than full-scene future representations. Its robustness under procedural task perturbations further indicates that delta tokens capture transferable task-relevant dynamics. These results suggest that efficient World-Action modeling need not reconstruct the future in full: predicting merely on what changes not only significantly reduces spatio-temporal redundancy and computational overhead, but also facilitates stronger policy learning.


\nocite{dumoulin2017style,strub2018multihopfilm,alayrac2022flamingo}
\bibliography{references}
\bibliographystyle{iclr2027_conference}

\clearpage
\appendix
\section*{\LARGE Appendix}
\section{Implementation Details}
\label{app:implementation}

\paragraph{Data and inputs.}
We use the two LIBERO RGB views (\texttt{agentview} and \texttt{eye\_in\_hand})
at $512\times512$ resolution. Four context frames per view are sampled at a
two-frame stride from 20\,Hz demonstrations. The policy receives a
nine-dimensional robot state and eight executed actions, and predicts 16
seven-dimensional actions.
DeltaWorld is adapted on the 6,500 demonstrations from LIBERO-Spatial, Object,
Goal, LIBERO-90, and LIBERO-10 with suite-balanced sampling. 
\paragraph{Architecture.}
Frozen DINOv3 ViT-B/16 produces a $32\times32$ patch grid per view, and
DeltaTok compresses each transition to one 768-dimensional token per view.
DeltaWorld and Action DiT each use 12 transformer blocks, width 768, and 12
attention heads. Current DINO patch anchors and predicted delta tokens receive
camera, temporal, and token-type embeddings after projection into the action
hidden space. Layer-wise KV caches for the Action DiT are reused across flow steps in the generation of one action chunk. 

\paragraph{Two-stage training.}
We adapt the Kinetics-pretrained DeltaWorld predictor while keeping DINOv3,
DeltaTok, and T5-Base frozen. Camera embeddings and zero-initialized gated
language modulation are trainable. Subsampling is applied---each delta step corresponds to two action steps. In the policy-training stage, the adapted DeltaWorld
is frozen. During this stage, DeltaWorld first samples eight future branches, then the Action DiT conditions on the minimum-error branch with probability $0.5$ and on a uniformly-sampled branch otherwise. 

\begin{table}[H]
\centering
\small
\renewcommand{\arraystretch}{0.96}
\caption{Training Hyperparameters.}
\label{tab:implementation_hyperparameters}
\begin{tabular}{lcc}
\toprule
\textbf{Setting} & \textbf{DeltaWorld adaptation} & \textbf{Action DiT training} \\
\midrule
Optimizer & AdamW & AdamW \\
Learning rate & $10^{-4}$ & $5\times10^{-5}$ \\
Weight decay & $0.4$ & $0.01$ \\
Warmup steps & 1,000 & 500 \\
Batch size / GPU & 32 & 2 \\
Gradient accumulation & 32 & 2 \\
Gradient clip norm & $0.5$ & $1.0$ \\
Future candidates & 256 (Best-of-Many) & 8 rollouts \\
Future horizon / view & 6 delta steps & 6 delta steps \\
\bottomrule
\end{tabular}
\end{table}

\paragraph{Inference and evaluation.}
At test time, DeltaWorld samples one trajectory without
oracle selection. Action DiT uses ten Euler steps, executes eight actions per
query, and temporally ensembles overlapping chunks with age-decay $0.1$.
Continuous end-effector actions are clipped to $[-1,1]$ and the gripper is
thresholded to $\{-1,+1\}$. Both in standard LIBERO and LIBERO-PRO we conduct 50 trials per task (500 per suite) and reports terminal success under three different random seeds.

\section{LIBERO-Plus Evaluation}
\label{app:libero_plus}
LIBERO-Plus evaluates robustness to controlled visual and embodiment distribution shifts \citep{fei2026liberoplus}. We report success rates under camera, robot, lighting, and layout perturbations, performing one trial per perturbation instance across all test suites.

\begin{table}[H]
\centering
\small
\renewcommand{\arraystretch}{0.96}
\caption{Success rates (\%) across selected LIBERO-Plus perturbation suites. Published baseline results are taken from the official benchmark leaderboard and the WAM robustness study \citep{fei2026liberoplus,zhang2026wamrobustness}.}
\label{tab:libero_plus}
\begin{tabular}{lcccc}
\toprule
\textbf{Scope / Method} & \textbf{Camera} & \textbf{Robot} & \textbf{Light} & \textbf{Layout} \\
\midrule
OpenVLA~\citep{kim2024openvla} & 0.8 & 3.5 & 8.1 & 28.5 \\
NORA~\citep{hung2025nora} & 2.2 & 37.0 & 45.7 & 62.1 \\
WorldVLA~\citep{cen2025worldvla} & 0.1 & 27.9 & 43.7 & 38.0 \\
UniVLA~\citep{bu2025univla} & 1.8 & 46.2 & 69.0 & 31.9 \\
$\pi_0$~\citep{black2024pi0} & 13.8 & 6.0 & \textbf{85.0} & \textbf{68.9} \\
$\pi_0$-FAST~\citep{pertsch2025fast} & 65.1 & 21.6 & 73.2 & 68.8 \\
Fast-WAM~\citep{yuan2026fastwam} & 16.4 & 44.5 & 78.2 & 60.7 \\
\midrule
\textbf{DeltaWAM (Average)} & \textbf{66.9} & 62.5 & \textbf{84.2} & 64.1\\
\bottomrule
\end{tabular}
\end{table}

\begin{figure}[!htbp]
    \centering
    \includegraphics[width=0.75\linewidth]{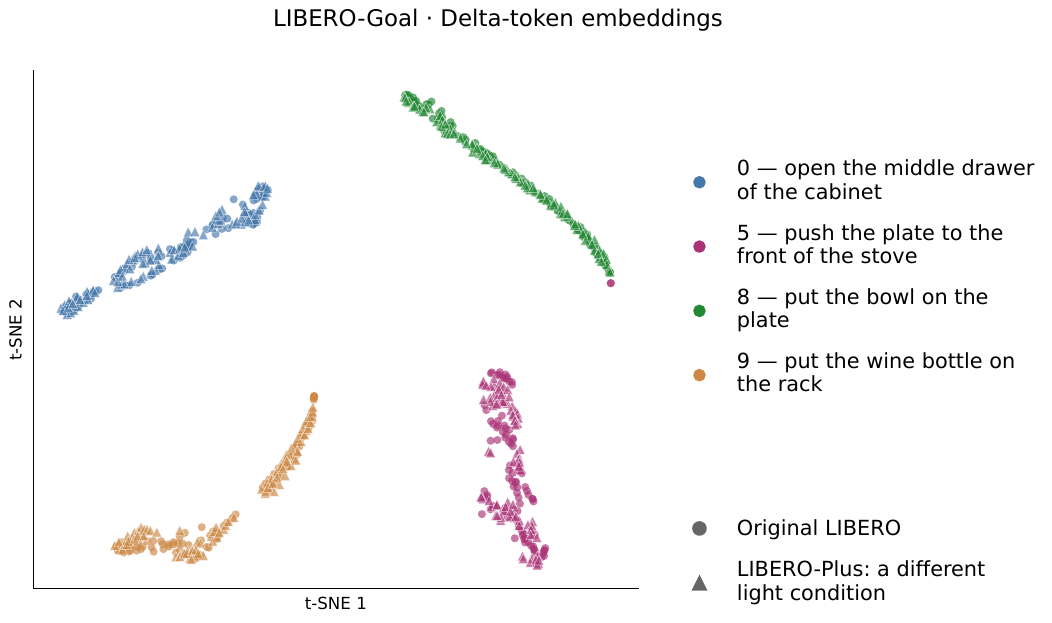}
    \caption{\textbf{Delta-token structure under LIBERO-Plus lighting variation.} LIBERO-Goal task embeddings under original light (circles) and a changed light condition (triangles).}
    \label{fig:libero_plus_light_tsne}
\end{figure}

Across suites, DeltaWAM achieves 66.9\%, 62.5\%, 84.2\%, and 64.1\% success under camera, robot, lighting, and layout perturbations, respectively. Figure~\ref{fig:libero_plus_light_tsne} shows that under lighting perturbations, the delta representations remain close to those of the original tasks. This provides a new perspective on improving robustness besides collecting more diverse data---we can make future representations more invariant to environmental perturbations. Delta tokens already partially show this property by focusing on dynamic changes.

\section{Additional Representation Analysis}
\label{app:representation_analysis}

\subsection{Analysis of DeltaWorld's Future Predictions}
\begin{figure}[!htbp]
    \centering
    \begin{minipage}[t]{0.49\linewidth}
        \centering
        \includegraphics[width=\linewidth]{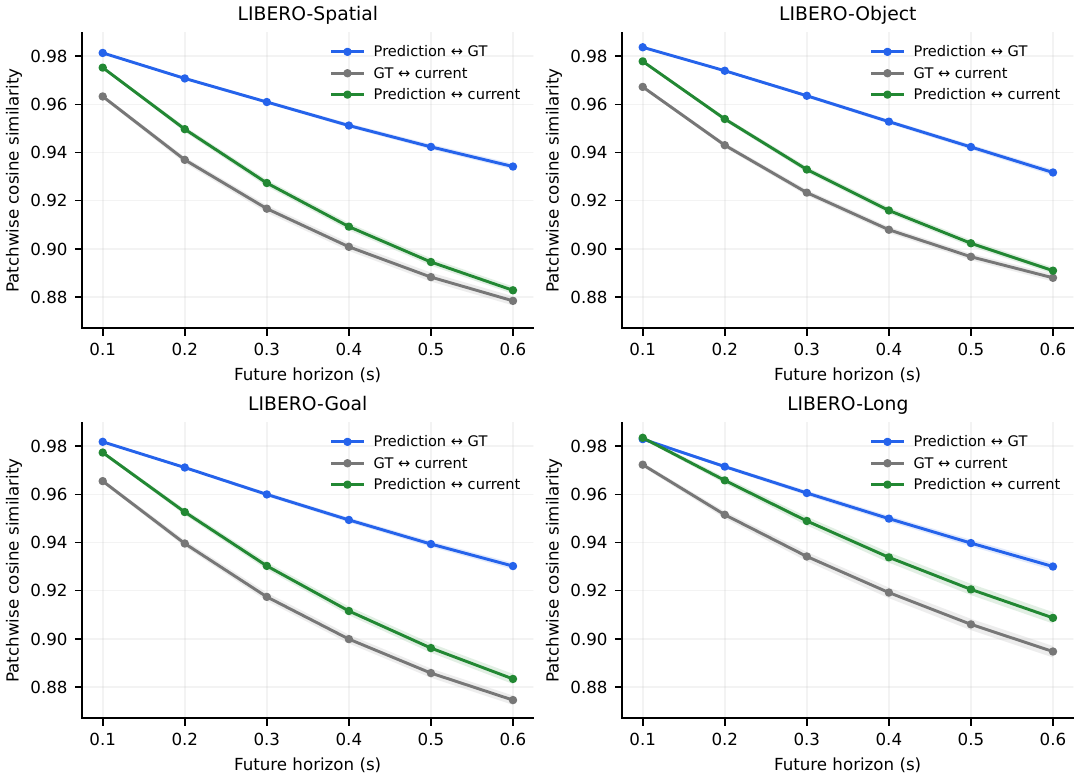}
        \caption{\textbf{Future-feature alignment across prediction horizons.} Patchwise DINO cosine similarity on training demonstrations compares predicted futures with ground truth (blue), ground truth with the current frame (gray), and predictions with the current frame (green). Six steps span 0.1--0.6\,s. }
        \label{fig:training_horizon_cosine}
    \end{minipage}
    \hfill
    \begin{minipage}[t]{0.49\linewidth}
        \centering
        \includegraphics[width=\linewidth]{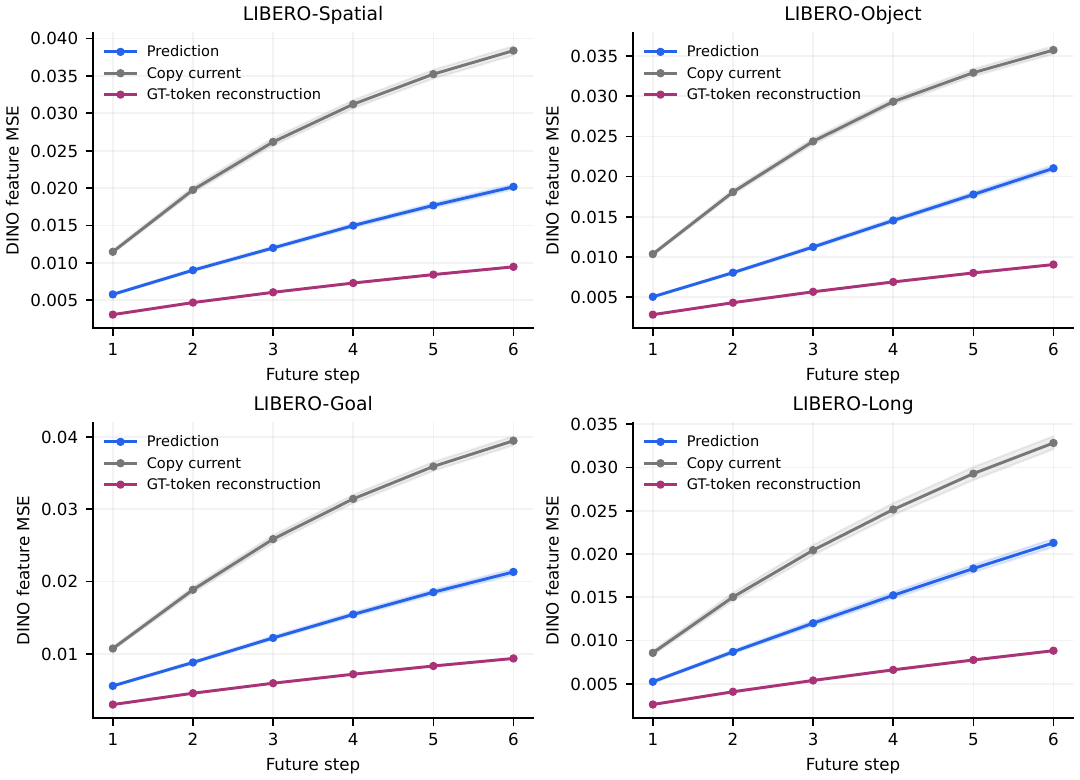}
        \caption{\textbf{Future prediction improves over copying the current state.} DINO feature MSE against actual future features for predicted delta-token rollouts (blue), a copy-current baseline (gray), and decoding delta tokens encoded from ground-truth futures (magenta). The latter is a reconstruction reference with access to the actual future.}
        \label{fig:training_horizon_mse}
    \end{minipage}
\end{figure}

We evaluate decoded predictions against future DINO features, using six future steps across all four suites. Figure~\ref{fig:training_horizon_cosine} shows that prediction--future similarity remains high as the horizon grows, while both predicted and observed futures become less similar to the current frame. Thus, the predicted representations capture temporal change rather than simply remaining near the initial observation. 

Figure~\ref{fig:training_horizon_mse} provides a complementary comparison with a static baseline: predicted features have lower MSE than copying the current frame at every displayed horizon in all four suites. Unavoidably, the MSE of DeltaWorld's predictions gradually increases as the future step becomes larger, which is a natural error accumulation for autoregressive predictions. Because of this accumulating error, training the policy with 6 future steps rather than 8 future steps(which correspond to the full horizon of one action chunk) actually gets better performance, as shown in table\ref{tab:future_horizon_success}.

\begin{table}[H]
\centering
\small
\caption{LIBERO success rates (\%) with six versus eight future steps.}
\label{tab:future_horizon_success}
\begin{tabular}{lccccc}
\toprule
\textbf{Future steps} & \textbf{Spatial} & \textbf{Object} & \textbf{Goal} & \textbf{Long} & \textbf{Average} \\
\midrule
Default (6) & 92.6 & 97.0 & 95.8 & 85.8 & 92.8 \\
Full horizon (8) & 88.8 & 95.4 & 94.6 & 81.2 & 90.0 \\
\bottomrule
\end{tabular}
\end{table}

\subsection{Task Structure in Delta-Token Embeddings}
\begin{figure}[!htbp]
    \centering
    \begin{minipage}[t]{0.49\linewidth}
        \centering
        \includegraphics[width=\linewidth]{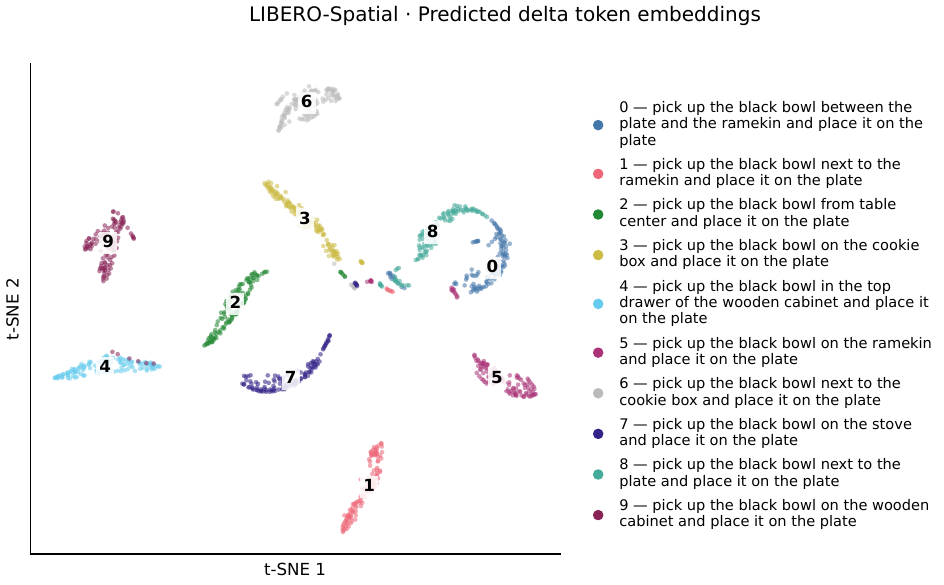}
        (a) LIBERO-Spatial
    \end{minipage}
    \hfill
    \begin{minipage}[t]{0.49\linewidth}
        \centering
        \includegraphics[width=\linewidth]{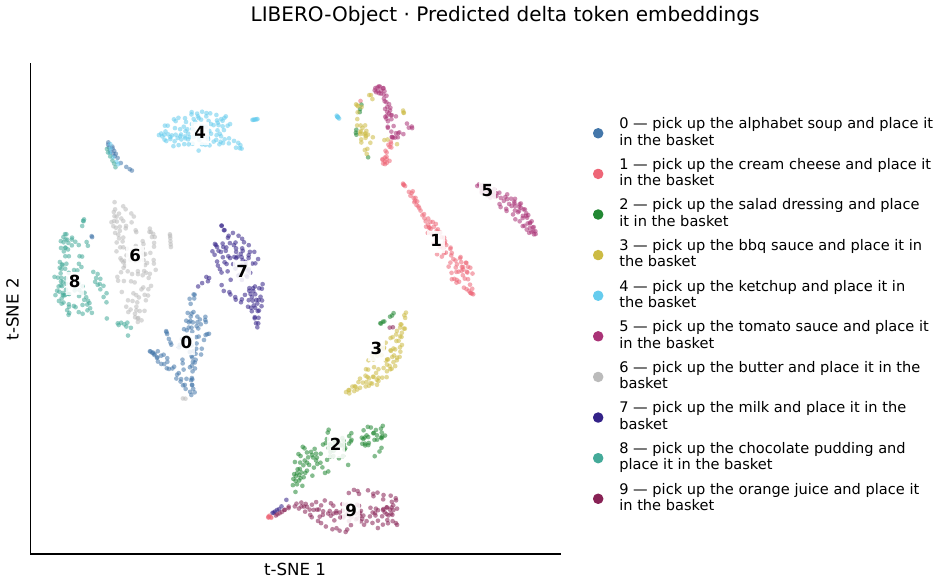}
        (b) LIBERO-Object
    \end{minipage}

    \vspace{0.5em}

    \begin{minipage}[t]{0.49\linewidth}
        \centering
        \includegraphics[width=\linewidth]{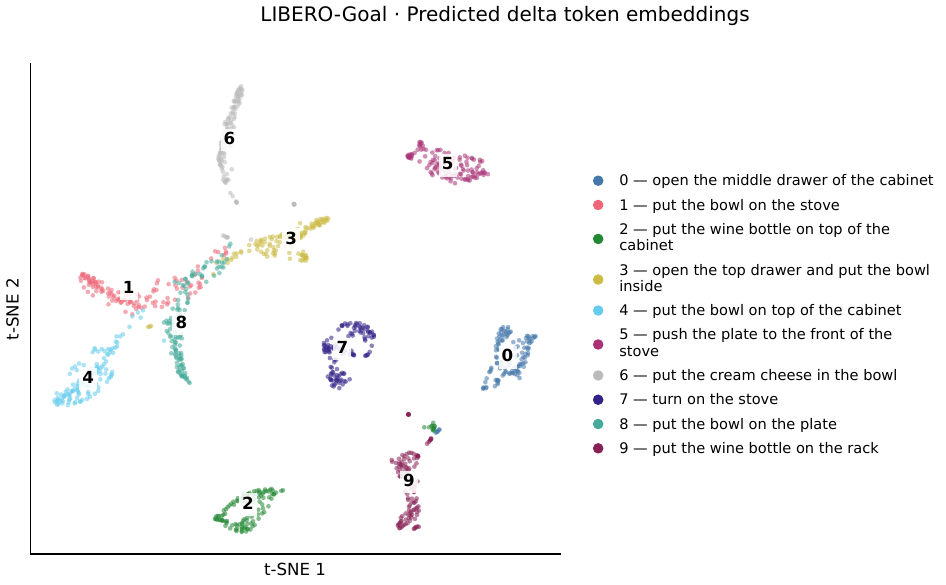}
        (c) LIBERO-Goal
    \end{minipage}
    \hfill
    \begin{minipage}[t]{0.49\linewidth}
        \centering
        \includegraphics[width=\linewidth]{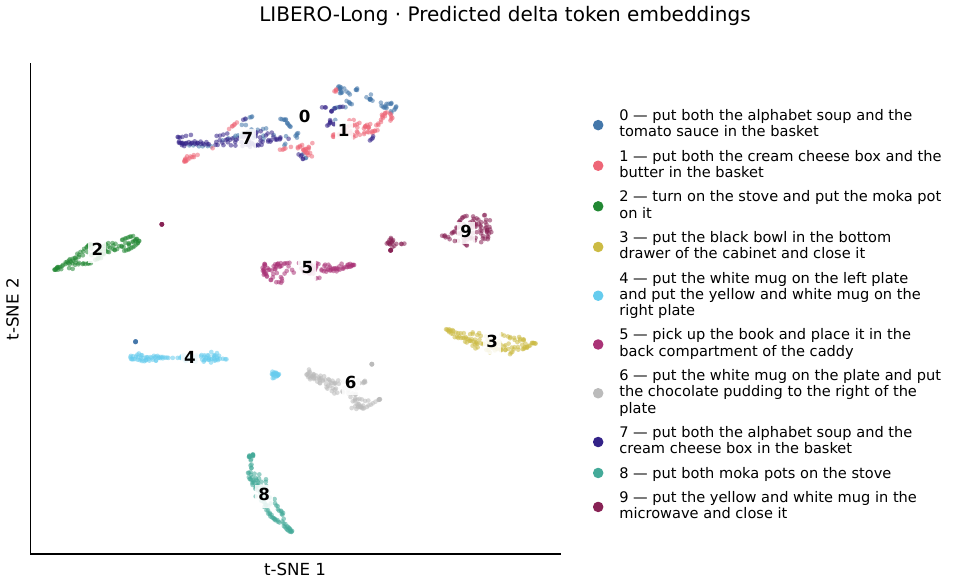}
        (d) LIBERO-10
    \end{minipage}
    \caption{\textbf{Task structure in predicted delta tokens across standard LIBERO suites.} Each point concatenates synchronized agent-view and wrist-view tokens at one future step, yielding one joint embedding per step. Samples come from middle-phase windows of collected rollouts.}
    \label{fig:delta_token_tsne_suites}
\end{figure}

Figure~\ref{fig:delta_token_tsne_suites} visualizes predicted delta tokens across the four standard LIBERO suites. The delta token embeddings of tasks that share the similar underlying semantics are generally closer. These qualitative projections are consistent with task-dependent future representations.

\subsection{Analysis of Action Attention to Predicted Future Tokens}
\label{app:action_attention}
\begin{figure}[!htbp]
    \centering
    \begin{minipage}[t]{0.495\linewidth}
        \centering
        \includegraphics[width=\linewidth]{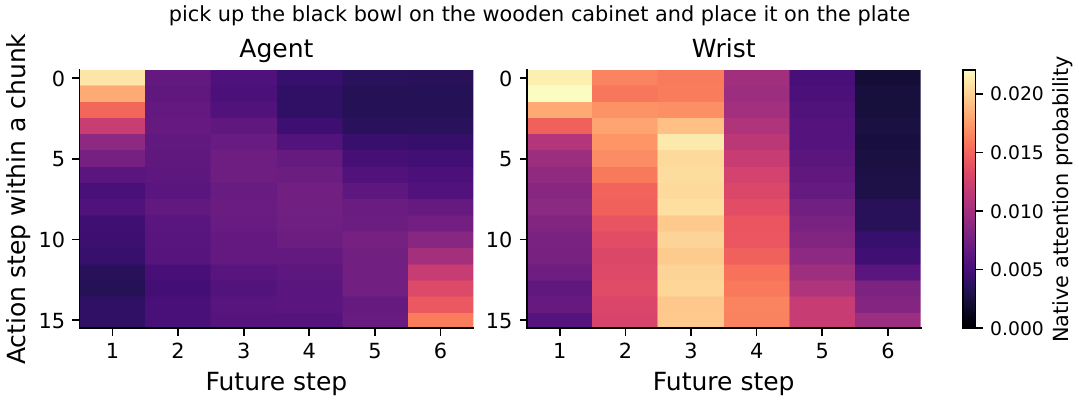}
        {\scriptsize (a) LIBERO-Spatial, task 9}
    \end{minipage}
    \hfill
    \begin{minipage}[t]{0.495\linewidth}
        \centering
        \includegraphics[width=\linewidth]{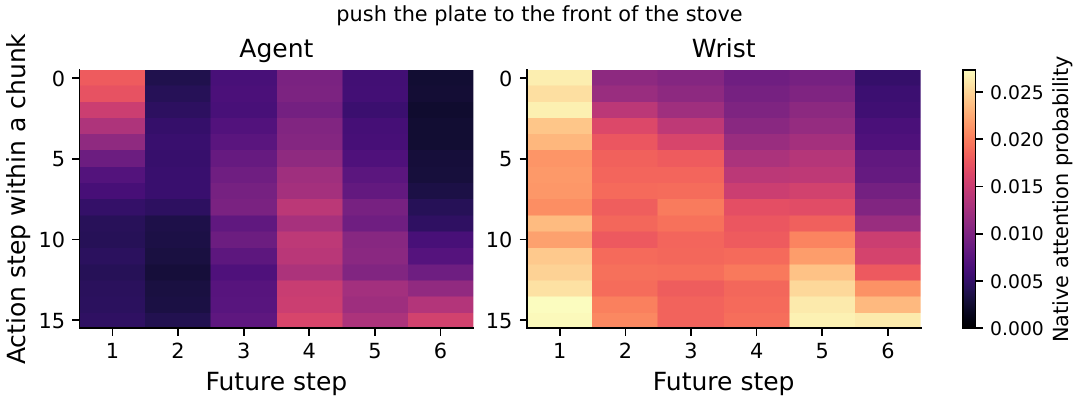}
        {\scriptsize (b) LIBERO-Goal, task 5}
    \end{minipage}

    \vspace{0.35em}

    \begin{minipage}[t]{0.495\linewidth}
        \centering
        \includegraphics[width=\linewidth]{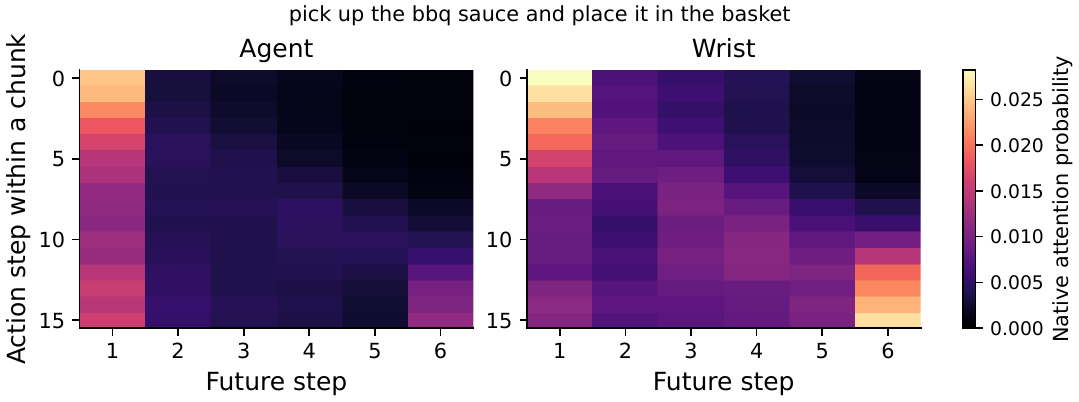}
        {\scriptsize (c) LIBERO-Object, task 3}
    \end{minipage}
    \hfill
    \begin{minipage}[t]{0.495\linewidth}
        \centering
        \includegraphics[width=\linewidth]{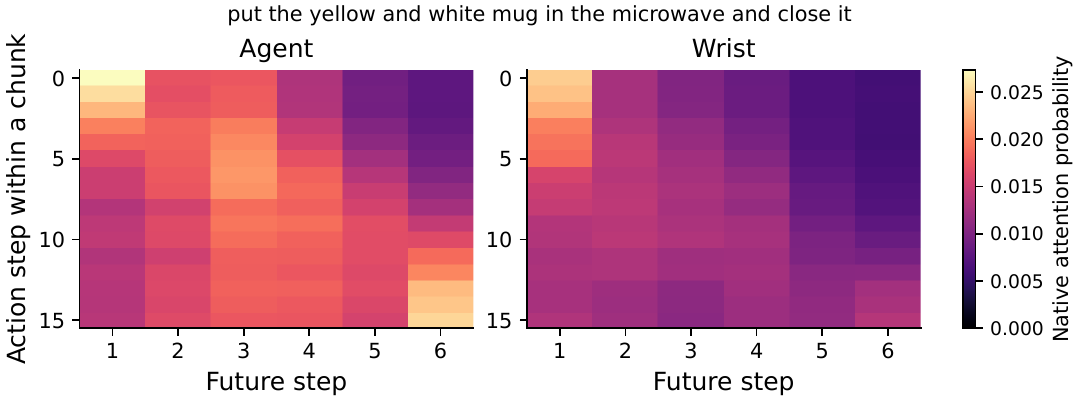}
        {\scriptsize (d) LIBERO-10, task 9}
    \end{minipage}
    \caption{\textbf{Action DiT attention to predicted future delta tokens.} Native cross-attention probabilities are averaged across Action DiT layers. Rows denote action positions within a generated chunk, columns denote predicted future steps, and agent and wrist tokens are shown separately. Probabilities remain normalized over the complete conditioning context, not only the displayed future tokens.}
    \label{fig:future_token_attention}
\end{figure}

To examine how the action expert uses the predicted sequence, we visualize its
native cross-attention probabilities from each action query to the future delta
tokens. Figure~\ref{fig:future_token_attention} shows representative examples
from all four standard LIBERO suites. Attention is non-uniform across both future
steps and action positions, and the patterns differ between agent and wrist
views. Early action queries often emphasize near-term tokens, while later queries
can shift weight toward more distant future steps.

\end{document}